\documentclass[conference]{IEEEtran}
\IEEEoverridecommandlockouts
\usepackage{cite}
\usepackage{amsmath,amssymb,amsfonts}
\usepackage{graphicx}
\usepackage{textcomp}
\usepackage{xcolor}
\usepackage{tikz}
\usepackage{subcaption}  
\usepackage[normalem]{ulem}
\usepackage{url} 
\usepackage{hyperref}

\usepackage{algorithm}
\usepackage{algorithmicx}
\usepackage{algpseudocode}

\usepackage{caption}

\newcommand{\nuc}{\newcommand}
\nuc{\clr}{\color{red}}
\nuc{\lab}{\label}
\nuc{\nn}{\nonumber}
\nuc{\cc}{\cite}
\nuc{\Sg}{\Sigma}
\nuc{\sg}{\sigma}

\usetikzlibrary{positioning}
\def\BibTeX{{\rm B\kern-.05em{\sc i\kern-.025em b}\kern-.08em
    T\kern-.1667em\lower.7ex\hbox{E}\kern-.125emX}}
\begin{document}

\title{3D Reconstruction via Incremental\\ Structure From Motion 
}

\author{\IEEEauthorblockN{Muhammad Zeeshan, Umer Zaki, Syed Ahmed Pasha}
\IEEEauthorblockA{Department of Electrical and Computer Engineering \\
Air University \\
Islamabad, Pakistan}
\and

\IEEEauthorblockN{Zaar Khizar}
\IEEEauthorblockA{Institut Pascal \\
Universit\'{e} Clermont Auvergne \\
Clermont-Ferrand, France}
}

\maketitle

\begin{abstract}
Accurate 3D reconstruction from unstructured image collections is a key requirement in applications such as robotics, mapping, and scene understanding. While global Structure from Motion (SfM) techniques rely on full image connectivity and can be sensitive to noise or missing data, incremental SfM offers a more flexible alternative. By progressively incorporating new views into the reconstruction, it enables the system to recover scene structure and camera motion even in sparse or partially overlapping datasets. In this paper, we present a detailed implementation of the incremental SfM pipeline, focusing on the consistency of geometric estimation and the effect of iterative refinement through bundle adjustment. We demonstrate the approach using a real dataset and assess reconstruction quality through reprojection error and camera trajectory coherence. The results support the practical utility of incremental SfM as a reliable method for sparse 3D reconstruction in visually structured environments.
\end{abstract}

\section{Introduction}
\label{sec:intro}

3D reconstruction is a fundamental task in computer vision that recovers the spatial structure of real world scenes from 2D images, transforming flat visuals into models with depth and geometry. Based on principles such as projective geometry, epipolar constraints, and image matching, 3D reconstruction has evolved from geometric methods to hybrid techniques combining classical and data-driven approaches \cc{HZi03, szeliski2022computer, pierson2017deep}. It finds applications in robotics \cite{lou2023slam}, medical imaging \cite{szeliski2022computer}, cultural heritage preservation \cite{xu2024real}, and smart city development \cite{wijayasinghe2024smart}, offering improved spatial measurements and scene understanding.

Traditional methods can be classified into geometric techniques such as triangulation \cite{HZi03}, volumetric fusion \cite{curless1996volumetric}, Structure from Motion (SfM) \cite{SFr16}, and Multi View Stereo (MVS) \cite{FHe15}. Triangulation estimates the 3D position of a point by intersecting rays from different viewpoints, with active and passive variants. Volumetric fusion integrates depth data from multiple views into a continuous 3D model. SfM detects and matches features across multiple images to estimate camera poses and generates a sparse 3D point cloud, while MVS refines this reconstruction using photometric consistency for dense depth maps. These methods are mathematically grounded and accurate under favorable conditions, such as good texture and wide camera baselines \cite{szeliski2022computer}.

The incremental SfM method starts with an initial image pair and progressively integrates new images by estimating their poses and triangulating 3D points \cite{SFr16}. This incremental approach is robust to noise and missing data, making it ideal for large scale reconstructions. Bundle adjustment refines camera parameters and 3D points by minimizing reprojection errors, enhancing consistency and accuracy \cite{triggs1999bundle}. A prominant example is COLMAP, an open-source SfM and MVS pipeline \cite{she2024refractive}. 

Recent developments have introduced hybrid and multi-camera SfM approaches that address scalability and robustness in challenging scenarios. AdaSfM combines coarse global SfM aided by IMU and encoder data with fine local incremental SfM to improve accuracy and efficiency in large-scale scenes \cite{chen2023adasfm}. MCSfM focuses on multi-camera systems, enabling automatic calibration and incremental reconstruction using rigid units and a two-stage bundle adjustment scheme \cite{cui2023mcsfm}. Line-based incremental SfM leverages geometric line features along with two observer strategies: a memoryless observer for real-time pose updates and a moving horizon observer that integrates a short history of measurements for improved stability\cite{mateus2021incremental}.

Recently, deep learning has emerged as an alternative to traditional methods, based on  techniques like convolutional neural networks for tasks such as depth estimation \cite{ranftl2020towards} and 3D scene understanding \cite{zhao2022monovit}. Models like MVSNet \cite{yao2018mvsnet} predict depth maps or voxel grids directly from images by learning appearance and geometry patterns. These methods often outperform classical techniques in textureless areas and occlusions, though they require large, labeled datasets and may not generalize well to dynamic or complex environments \cite{zhang2020review}.

In this paper, we revisit classical geometric methods by presenting a modular, interpretable incremental SfM pipeline. We demonstrate that traditional techniques remain effective for accurate and consistent 3D reconstruction from unordered image sets. The pipeline is reproducible and well-suited for both research and practical use. Through reprojection error analysis and comparison with COLMAP, we show that our method achieves competitive accuracy while offering greater transparency and flexibility.

The paper is structured as follows: Section~{II} provides some background on tools needed for incremental SfM, and Section~{III} discusses the incremental SfM pipline. Experimental results in presented in Section~{IV}. We offer some conclusions in Section~{V}.

\section{Background}
\label{sec:bg}

This section briefly reviews key mathematical tools used in the incremental Structure from Motion (SfM) pipeline. We begin by introducing some notation.

Given $X\in\mathbb{R}^{n\times m}$, $x=\text{vec}(X)\in\mathbb{R}^{nm}$ is the column stacking operation. $\otimes$ and $*$ denote the Kronecker product and convolution operators respectively. $I_n$ and $0_{n\times m}$ are the $n\times n$ identity matrix and  $n\times m$ matrix of all zeros respectively. Given $x=(x_1,x_2,x_3)^T \in\mathbb{R}^3$,  $\|x\|=\sqrt{x^Tx}$, and $[x]_{\times}$ is the corresponding skew-symmetric matrix, 
\[
[x]_{\times}= \begin{bmatrix} 
0 & -x_3 & x_2 \\ 
x_3 & 0 & -x_1 \\
-x_2 & x_1 & 0 \\
\end{bmatrix}.
\]

\subsection{Direct Linear Transform}
\label{sec:dlt}
The Direct Linear Transformation (DLT) is commonly used to compute the camera projection matrix that maps 3D world points to their corresponding 2D image projections, based on known point correspondences.

A 3D point \( P_i = (X_i, Y_i, Z_i)^T \) is related to its 2D projection \( p_i = (x_i, y_i)^T \), via the camera projection matrix \( M \in \mathbb{R}^{3 \times 4} \) in homogeneous coordinates as
\begin{align}
\begin{bmatrix}
p_i \\
1
\end{bmatrix} &= M 
\begin{bmatrix}
P_i \\
1
\end{bmatrix}, \label{eq:proj} \\
&= \left( \begin{bmatrix}
P_i^T & 1 
\end{bmatrix} \otimes I_3 \right) m \nn
\end{align}
where $m = \text{vec}(M) \in \mathbb{R}^{12}$. 

Let $n=\#$ (known) point correspondences and   
\[
A_i = \left( \begin{bmatrix}
P_i^T & 1 
\end{bmatrix} \otimes I_3 \right), \quad b_i = \begin{bmatrix}
p_i^T & 1 
\end{bmatrix}^T, \quad i=1,2,...,n,
\]
we can assemble the linear system $A m = b$, with $A \in \mathbb{R}^{3n \times 12}$ and $b \in \mathbb{R}^{3n}$ given by
\[
A= \begin{bmatrix}
A_1^T & A_2^T & \cdots & A_n^T 
\end{bmatrix}^T, \quad b= \begin{bmatrix}
b_1^T & b_2^T & \cdots & b_n^T 
\end{bmatrix}^T. 
\]

\textit{Remark 1:} A minimum of $n=4$ point correspondences is needed to ensure the linear system is not ill-conditioned. 

\medskip

To improve stability in the presence of noise, the linear system is solved via the singular value decomposition (SVD) \cc{golub2013matrix}. 
Take the SVD, \( A = U_A\Sg_AV_A^T \), the solution \( m \) is the last column of \( V_A \), corresponding to the smallest singular value. Reshape $m$ to construct the projection matrix $M$. 

To extract the intrinsic matrix $K$, rotation matrix ${\cal R}$, and translation vector $t$ from $M = \begin{bmatrix}
H & h 
\end{bmatrix}$ where $H = K{\cal R}\in\mathbb{R}^{3\times 3}$, the QR decomposition \cite{golub2013matrix} can be used.
Let $H^{-1} = QR$, then $H = R^{-1} Q^T$, which gives $K = R^{-1}, {\cal R} = Q^T$, and $t = -H^{-1} h$.

\subsection{Scale-Invariant Feature Transform}
\label{sec:sift}
Scale-Invariant Feature Transform (SIFT) \cite{Low04} is a widely used algorithm to detect and describe distinctive local features in images. It identifies keypoints that are invariant to scale, rotation, and partially invariant to affine transformations and changes in illumination. This robustness makes SIFT particularly effective for tasks such as object recognition, image stitching, and 3D reconstruction. The algorithm operates in four main stages: scale-space extrema detection, keypoint localization, orientation assignment, and descriptor generation.

To detect features across different scales, a \textit{scale space} is constructed by applying Gaussian blur to the input image using Gaussian kernels with progressively increasing standard deviation. The Difference of Gaussians (DoG) is then computed by subtracting adjacent Gaussian-blurred images
\[
D(x, y, \sigma) = L(x, y, k\sigma) - L(x, y, \sigma),
\]
where \( L(x, y, \sigma) = I(x, y) * G(x, y, \sigma) \) is the image convolved with a Gaussian filter of scale \( \sigma \), and \( k > 1 \) is a constant multiplicative factor. Keypoints are detected as local extrema in the DoG images across both spatial and scale dimensions.

To build the DoG pyramid, Gaussian blurred images are generated by increasing \(\sigma\) and adjacent images are subtracted to highlight intensity changes \cite{marr1980theory}. Keypoints are detected via non maximum suppression by comparing each pixel with its 26 neighbors across adjacent scales.

To ensure rotation invariance, an orientation is assigned to each keypoint based on local image gradients, typically obtained using the Sobel filters \cite{freeman1991design}. 
The dominant orientation is selected from a histogram of gradient orientations in the local neighborhood of the keypoint.

Finally, a $16 \times 16$ region around each keypoint is divided into $4 \times 4$ cells, and in each cell an 8 bin histogram of gradient directions is computed. These histograms are concatenated to form a 128 dimensional feature vector, and normalized to reduce the effects of changes in illumination.


\section{Methodology}

\begin{figure*}[t]
    \centering
    \includegraphics[width=\textwidth]{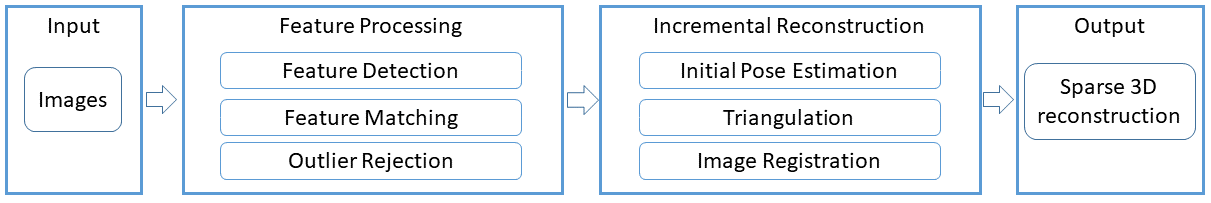}
    \caption{Incremental SfM Pipeline.}
    \label{fig:BD}
\end{figure*}

The incremental Structure from Motion (SfM) pipeline (see Fig.~\ref{fig:BD}), begins with intrinsic camera calibration, followed by feature detection and matching across image pairs. An initial image pair is selected, and the relative pose is estimated using epipolar geometry. A sparse 3D point cloud is then computed through triangulation. Subsequent images are registered incrementally by estimating their poses via 2D--3D correspondences. Additional 3D points are triangulated from observations in the new views. At each stage, bundle adjustment is applied to jointly refine camera parameters and 3D point locations, minimizing reprojection error and maintaining geometric consistency.

Camera calibration was used to estimate the intrinsic matrix $K$, which contains the camera’s focal lengths $(f_x, f_y)$ and principal points $(c_x, c_y)$ \cc{HZi03}. The projection of a 3D world point $P_i$ to a 2D image point $p_i$ is given by \eqref{eq:proj}, where the camera projection matrix $M$ was estimated using the DLT algorithm outlined in \autoref{sec:dlt}.

\subsection{Feature Detection and Matching}

Keypoints were extracted using the SIFT algorithm (see \autoref{sec:sift}), which ensures invariance to scale and rotation and provides robust 128-dimensional descriptors.

Feature matching was performed using brute force matching based on the Euclidean distance. To eliminate outliers, we employed the RANSAC algorithm \cc{MJR22} in combination with the normalized 8-point algorithm \cc{Har97}. The 8-point algorithm is a linear method to estimate the fundamental matrix $\mathbf{F}$, which encapsulates the epipolar geometry between two views. 

Given a set of 8 or more corresponding points \( p_L = (x_L, y_L)^T \) and \( p_R = (x_R, y_R)^T \), where \( p_L \) and \( p_R \) are homogeneous coordinates in the left and right images respectively, the fundamental matrix \( \mathbf{F} \) satisfies the epipolar constraint~\cite{HZi03},
\begin{align*}
\begin{bmatrix} p_L \\ 1 \end{bmatrix}^T \mathbf{F} \begin{bmatrix} p_R \\ 1 \end{bmatrix} = 0 \\
\equiv 
\left( \begin{bmatrix} p_R^T & 1 \end{bmatrix} \otimes  \begin{bmatrix} p_L^T & 1 \end{bmatrix} \right) f = 0 
\end{align*}
with $f=\text{vec}(\mathbf{F})$.

This is a homogeneous linear system \( Gf = 0 \), with 
\[
G = \Big[ \begin{bmatrix} p_R^T & 1 \end{bmatrix} \otimes  \begin{bmatrix} p_L^T & 1 \end{bmatrix} \Big ],
\]
which can be solved using the SVD, where the solution is the singular vector corresponding to the smallest singular value.

Since the rank of matrix $\mathbf{F}$ is  $2$, the smallest singular value of $\mathbf{F}$ is set to zero, and the matrix is reconstructed as,
\[
\textbf{F} = U_F \Sg_F V_F^T, \quad 
\Sg_F = \text{diag}(\sg_1, \sg_2, 0).
\]

To improve numerical stability in the estimation of \( \mathbf{F} \), a normalization step is applied to the input points from each image \cc{Har97}. Specifically, the image coordinates are translated so that their centroid lies at the origin and scaled such that their average distance from the origin \(= \sqrt{2} \). For a set of \( n \) points \( \{(x_i, y_i)\}_{i=1}^n \), with mean \( (\bar{x}, \bar{y}) \), the average distance from the centroid is
\[
\bar{d} = \frac{1}{n} \sum_{i=1}^n \sqrt{(x_i - \bar{x})^2 + (y_i - \bar{y})^2}.
\]
The resulting similarity transformation matrix is
\[
T = \begin{bmatrix}
s & 0 & -s\bar{x} \\
0 & s & -s\bar{y} \\
0 & 0 & 1
\end{bmatrix}, \quad s = \frac{\sqrt{2}}{\bar{d}}.
\]

\textit{Remark 2:} The target value of \( \sqrt{2} \) is selected to bring the coordinate values to a comparable numerical scale, typically of order one, thereby improving the conditioning of the equations used in the estimation of \( \mathbf{F} \). Even when \( \bar{d} = \sqrt{2} \) and the scaling factor \( s = 1 \), the translation component of the normalization still plays a critical role by centering the data at the origin, which further contributes to numerical stability.

\medskip

Let \( T_1 \) and \( T_2 \) denote the normalization matrices for the point sets in the first and second images, respectively. After estimating the fundamental matrix \(\bar{\mathbf{F}} \), in the normalized coordinates, the unnormalized matrix $\mathbf{F}$ is recovered by applying the inverse normalization transformations as 
\[
\mathbf{F} = T_2^T \bar{\mathbf{F}} T_1.
\]

\subsection{Camera Pose Estimation}

Next we estimated the relative camera poses. This entails recovering the rotation and translation that define the spatial relationship between the views. Pose estimation is based on the essential matrix $\mathbf{E}$, which encodes the epipolar geometry between two images given the camera calibration matrix. For a pair of normalized corresponding points \( p_L \) and \( p_R \), the epipolar constraint is~\cite{HZi03},
\[
p_R^T \mathbf{E} p_L = 0.
\]

If the fundamental matrix \( \mathbf{F} \) is known, the essential matrix $\mathbf{E}$ can be computed using the intrinsic calibration matrices \( K_L \) and \( K_R \) of the two cameras as \cite{HZi03},
\[
\mathbf{E} = K_R^T \mathbf{F} K_L.
\]
For $K_R= K_L= K$, we have, 
\begin{equation}
\textbf{E} = K^T \textbf{F} K. \lab{e:E}
\end{equation}
This transformation maps pixel coordinates into normalized image coordinates, enabling pose estimation in calibrated space.

The matrix $\mathbf{E}$ can be further decomposed as 
\[
\textbf{E} = [t]_\times {\cal R},
\]
where \( [t]_\times \) is the skew-symmetric matrix of the translation vector \( t \), i.e.,
\[
[t]_\times = \begin{bmatrix}
0 & -t_z & t_y \\
t_z & 0 & -t_x \\
-t_y & t_x & 0
\end{bmatrix}.
\]

Taking the SVD of $\mathbf{E}$, among the four possible decompositions of \( {\cal R} \) and \( t \), the physically meaningful solution was selected using the cheirality condition~\cite{nister2004efficient}, ensuring that the reconstructed 3D points lie in front of both cameras.

With \( {\cal R} \) and \( t \) determined, the camera projection matrices were obtained as,
\[
M_L = [I_3\ |\ 0_{3\times 1}], \quad M_R = [{\cal R}\ |\ t],
\]
which are used in the triangulation to estimate the 3D structure.

The (initial) reconstruction steps outlined above form the basis for the incremental SfM pipeline, which incrementally registers new views, estimates their poses, triangulates additional points, and refines all parameters via bundle adjustment \cite{SFr16}.

\subsection{Triangulation}

Once the corresponding feature points were identified across multiple views, triangulation was employed to estimate the 3D coordinates of these points in space. This procedure involves computing the intersection of the back projected rays originating from each camera center through the respective 2D image points, ideally converging at a single 3D location\cite{HZi03}.

Let \( P = (X, Y, Z)^T \) be the 3D point in space to be reconstructed and the corresponding image coordinates in homogeneous form be \( \tilde{p}_L = (x_L, y_L, 1)^T \) and \( \tilde{p}_R = (x_R, y_R, 1)^T \) via the projection matrices \( M_L \) and \( M_R \), respectively, i.e.,
\[
\begin{bmatrix}
\tilde{p}_L \\
\tilde{p}_R \\
\end{bmatrix} =
\begin{bmatrix}
M_L \\
M_R \\
\end{bmatrix} \tilde{P} 
\]
where \( \tilde{P} = (P^T, 1)^T \) is the homogeneous representation of $P$.

To enforce the 3D point lies along the line of sight corresponding to each image observation, we imposed,
\[
\begin{bmatrix}
\tilde{p}_L \times (M_L \tilde{P}) \\
\tilde{p}_R \times (M_R \tilde{P})
\end{bmatrix} = 0,
\]
which led to the homogeneous linear system of equations,
\[
L\tilde{P} = 0
\]
with $L = \begin{bmatrix} L_L^T & L_R^T \end{bmatrix}^T$ and
\[
L_L = [\tilde{p}_L]_\times M_L, \quad
L_R = [\tilde{p}_R]_\times M_R.
\]

The solution to the linear system was obtained via the SVD of  \( L \), i.e., $L = U_L \Sg_L V_L^T$. Then, the triangulated 3D point (in homogeneous coordinates), \( \tilde{P} \) is the last column of matrix \( V_L \).

\subsection{Bundle Adjustment}
\label{sec:BA}
To refine the camera poses and 3D structure, we performed bundle adjustment, which minimizes the reprojection error across all observations via the optimization \cc{SFr16},
\[
\min_{\{M_j\}, \{P_i\}} \ \sum_{i,j} \| p_{ij} - \pi(M_j, P_i) \|^2,
\]
where \( p_{ij} \) is the observed 2D image coordinates of 3D point \( P_i \) in image \( j \), and \( \pi(M_j, P_i) \) is the projection of point \( P_i \) in image \( j \) using the camera matrix \( M_j \). A variant of the Levenberg–Marquardt algorithm, specifically Trust Region Reflective \cite{NWr99}, was employed for the nonlinear optimization to ensure globally consistent and accurate 3D reconstruction.

\subsection{Incremental Structure from Motion (SfM)}

In the incremental SfM pipeline, the reconstruction process begins by selecting an appropriate initial image pair to establish the global coordinate frame and initialize the 3D structure. This selection is guided by two primary criteria: the number of matched feature correspondences and the geometric diversity between the camera viewpoints. In particular, image pairs with a high number of inlier matches and sufficient spatial separation, commonly referred to as the baseline, are preferred. The baseline is defined as the Euclidean distance between the two camera centers, \( \mathbf{C}_1 \) and \( \mathbf{C}_2 \).

Assuming the first camera is positioned at the origin of the world coordinate system, i.e., \( \mathbf{C}_1 = 0 \), the relative pose of the second camera can be recovered by decomposing the essential matrix \( \mathbf{E} \). Since all the images are captured by the same calibrated camera moving through space, the intrinsic parameters remain constant across views. Thus, \( K_L = K_R = K \) and the decomposition \eqref{e:E} yields the relative rotation and translation (up to scale) between the views, provided that the camera intrinsics are known. Then, the second camera center
\begin{equation}
\mathbf{C}_2 = -\mathcal{R}^\top t \label{e:C2}
\end{equation}
and the baseline \(= \|t\| \). 

A longer baseline improves the accuracy of 3D triangulation. The quality of triangulation also depends on the angle \( \theta \) between the viewing rays from both cameras to a 3D point \( P \) with
\[
\cos\theta = \frac{(P - \mathbf{C}_1)^T (P - \mathbf{C}_2)}{\|P - \mathbf{C}_1\|  \|P - \mathbf{C}_2\|},
\]
where a larger angle \( \theta \) gives better triangulation geometry. The correct configuration of ${\cal R}$ and \( t \) is selected by enforcing the cheirality condition  \cite{nister2004efficient}.

Once the initial pair was processed, additional images were incrementally added using the Perspective-n-Point (PnP) algorithm\cite{lu2018review}. This algorithm estimates the camera pose for a new image using known 3D points \( P_i \in \mathbb{R}^3 \) and their corresponding 2D image projections \( p_i \in \mathbb{R}^2 \). The camera projection matrix \begin{equation} 
M = [{\cal R}\ |\ t] \lab{e:M} 
\end{equation}
maps 3D points to 2D points. The reprojection of each 3D point $p_i = \pi(M, P_i)$, for $i = 1,2, \dots, n$, where \( \pi(\cdot,\cdot) \) is the perspective division to obtain the pixel coordinates. The PnP algorithm minimizes the total reprojection (squared) error, 
\[
\min_{{\cal R}, t} \ \sum_{i=1}^{n} \| p_i - \pi(M ,P_i) \|^2.
\]

Efficient solutions such as EPnP \cite{lepetit2009ep} are used in combination with RANSAC to handle outliers. Once the pose of the new image was estimated, it was added to the reconstruction, and new 3D points were triangulated using matches with previously registered images. These points were integrated in the global model. Finally, the entire structure and camera poses were refined using bundle adjustment (see \autoref{sec:BA}).
The selection of the next image to register is guided by visibility and overlap. Images that observe a large number of already triangulated 3D points, allowing more 2D–3D correspondences were prioritized. This strategy, (aka greedy view selection) ensured robust PnP pose estimation and gradual, stable reconstruction expansion. The incremental SfM algorithm is summarized in Algorithm \ref{a:1}.

\begin{algorithm}
\caption{Incremental SfM} \lab{a:1}
\begin{algorithmic}[1]
%
\Procedure{DetectFeatures}{${\cal I}$}\Comment{Image collection ${\cal I}=\{I_i\}_1^N$}
  \State Convert $I_i$ for $i=1,...,N$ to grayscale
  \State Detect keypoints $\{p_j\}_1^{n_i}$ for $i=1,...,N$ using SIFT
  \State Assemble descriptors $D_i=\{d_j\}_1^{n_i}$ for $i=1,...,N$ 
\EndProcedure
\Procedure{MatchFeatures}{${\cal D}$}\Comment{Descriptor collection ${\cal D}=\{D_1,...,D_N\}$}
  \For{$(I_i, I_j)$}\Comment{$i=1,...,N-1, j>i$}
    \State Match descriptors $D_i, D_j$ to get correspondences ${\cal M}_{ij}$
    \State Filter correspondences using Lowe's ratio test
    \State Estimate essential matrix ${\bf E}_{ij}$ using RANSAC
  \EndFor
\EndProcedure
\State Select pair $(I_a, I_b)$ with maximum inlier correspondences
\State Estimate relative pose $[{\cal R}|t]$ from ${\bf E}_{ab}$
\State Triangulate initial 3D points $\{P_i\}$ from ${\cal M}_{ab}$
\State Initialize camera pose: $[{\cal R}_a|t_a] = [I_3|0_{3\times 1}]$, $[{\cal R}_b|t_b] = [{\cal R}|t]$
\State Add triangulated points to the point cloud $\mathcal{P}$
\Procedure{EstimatePose}{${\cal I}$}
  \While{unregistered images remain}
    \State Select $I_k$ with sufficient 2D--3D correspondences
    \State Estimate pose $[{\cal R}_k|t_k]$ using PnP with RANSAC
    \State Add $[\mathcal{R}_k|t_k]$ to pose graph
    \State Triangulate new 3D points with previously registered views
    \State Add valid points to ${\cal P}$
  \EndWhile
\EndProcedure
\Procedure{BundleAdjustment}{$\{[{\cal R}_i|t_i]\}_1^N,{\cal P}$}
  \State Jointly optimize all camera poses $\{[{\cal R}_i|t_i]\}$ for $i=1,...,N$, and 3D points ${\cal P}$ to minimize reprojection error
\EndProcedure
\end{algorithmic}
\end{algorithm}

\section{Experimental Results}
\label{sec:Res}

\begin{figure}[t] 
    \centering
    \begin{subfigure}[b]{0.15\textwidth}
        \includegraphics[width=\textwidth]{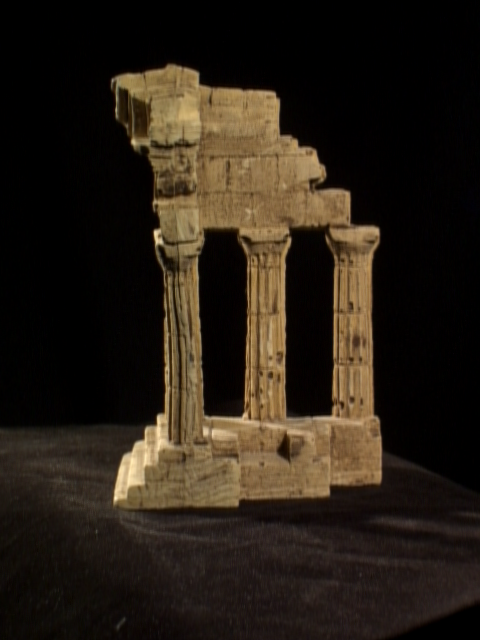}
        \caption{Frame 0}
    \end{subfigure}
    \begin{subfigure}[b]{0.15\textwidth}
        \includegraphics[width=\textwidth]{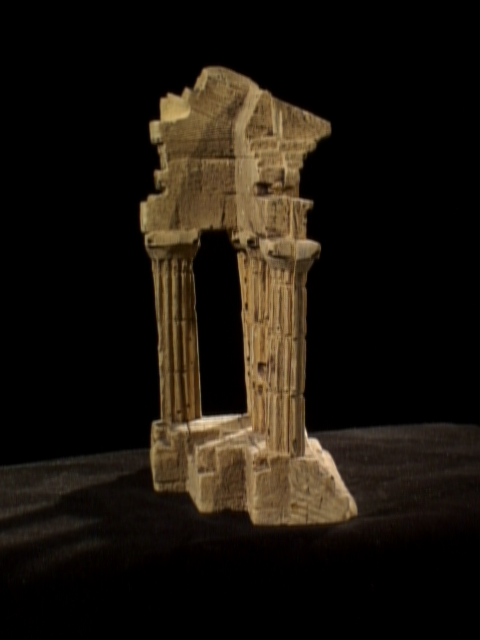}
        \caption{Frame 9}
    \end{subfigure}
    \begin{subfigure}[b]{0.15\textwidth}
        \includegraphics[width=\textwidth]{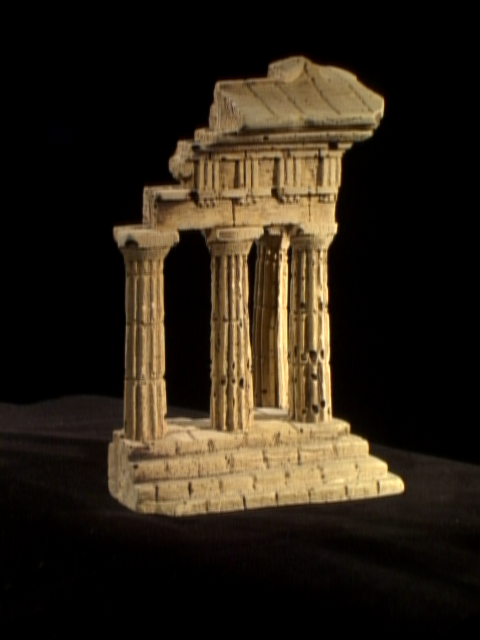}
        \caption{Frame 18}
    \end{subfigure}
    \begin{subfigure}[b]{0.15\textwidth}
        \includegraphics[width=\textwidth]{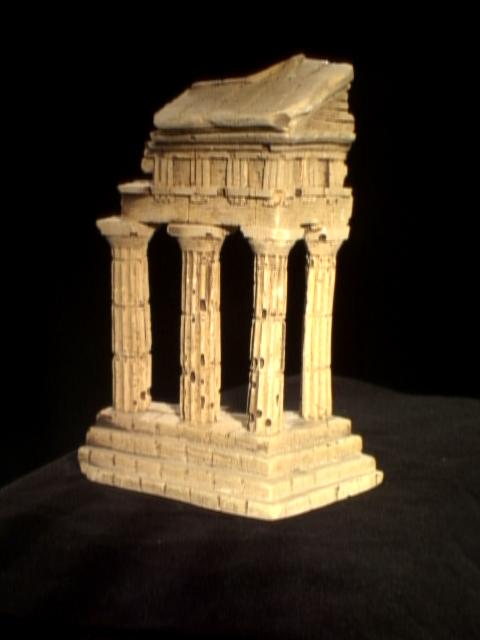}
        \caption{Frame 27}
    \end{subfigure}
    \begin{subfigure}[b]{0.15\textwidth}
        \includegraphics[width=\textwidth]{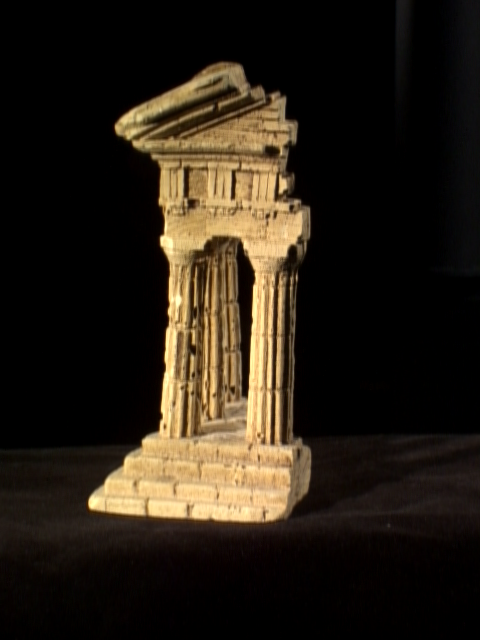}
        \caption{Frame 36}
    \end{subfigure}
    \begin{subfigure}[b]{0.15\textwidth}
        \includegraphics[width=\textwidth]{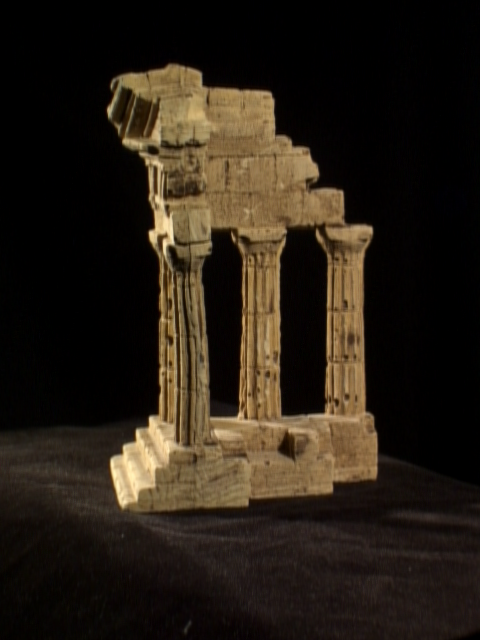}
        \caption{Frame 45}
    \end{subfigure}

    \caption{Temple Ring Dataset: Sample frames showing different viewpoints.}
    \label{fig:input_images_matrix}
\end{figure}

This section demonstrates the results of the incremental SfM pipeline\footnote{https://github.com/zaarAli/i-sfm} for the Temple Ring dataset\footnote{ \url{https://vision.middlebury.edu/mview/data/data/templeRing.zip}} which contains 47 high resolution images. Fig.~\ref{fig:input_images_matrix} shows selected images from the dataset that depict different viewpoints.

\begin{figure}[t] 
    \centering
    \begin{subfigure}[b]{0.22\textwidth}
        \centering
        \includegraphics[width=\textwidth]{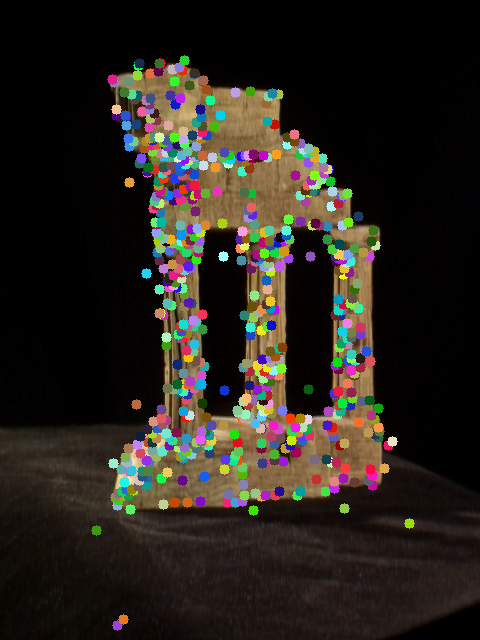}
        \caption{Keypoints in Frame 0}
    \end{subfigure}
    \hspace{1em}
    \begin{subfigure}[b]{0.22\textwidth}
        \centering
        \includegraphics[width=\textwidth]{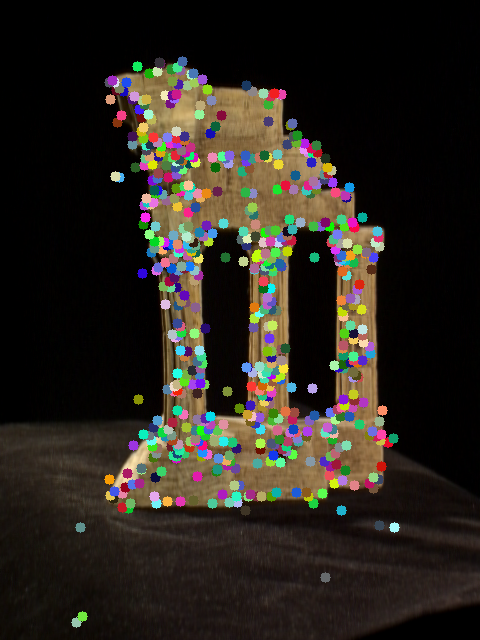}
        \caption{Keypoints in Frame 1}
    \end{subfigure}
    
    \vspace{1em}
    
    \begin{subfigure}[b]{0.48\textwidth}
        \centering
        \includegraphics[width=\textwidth]{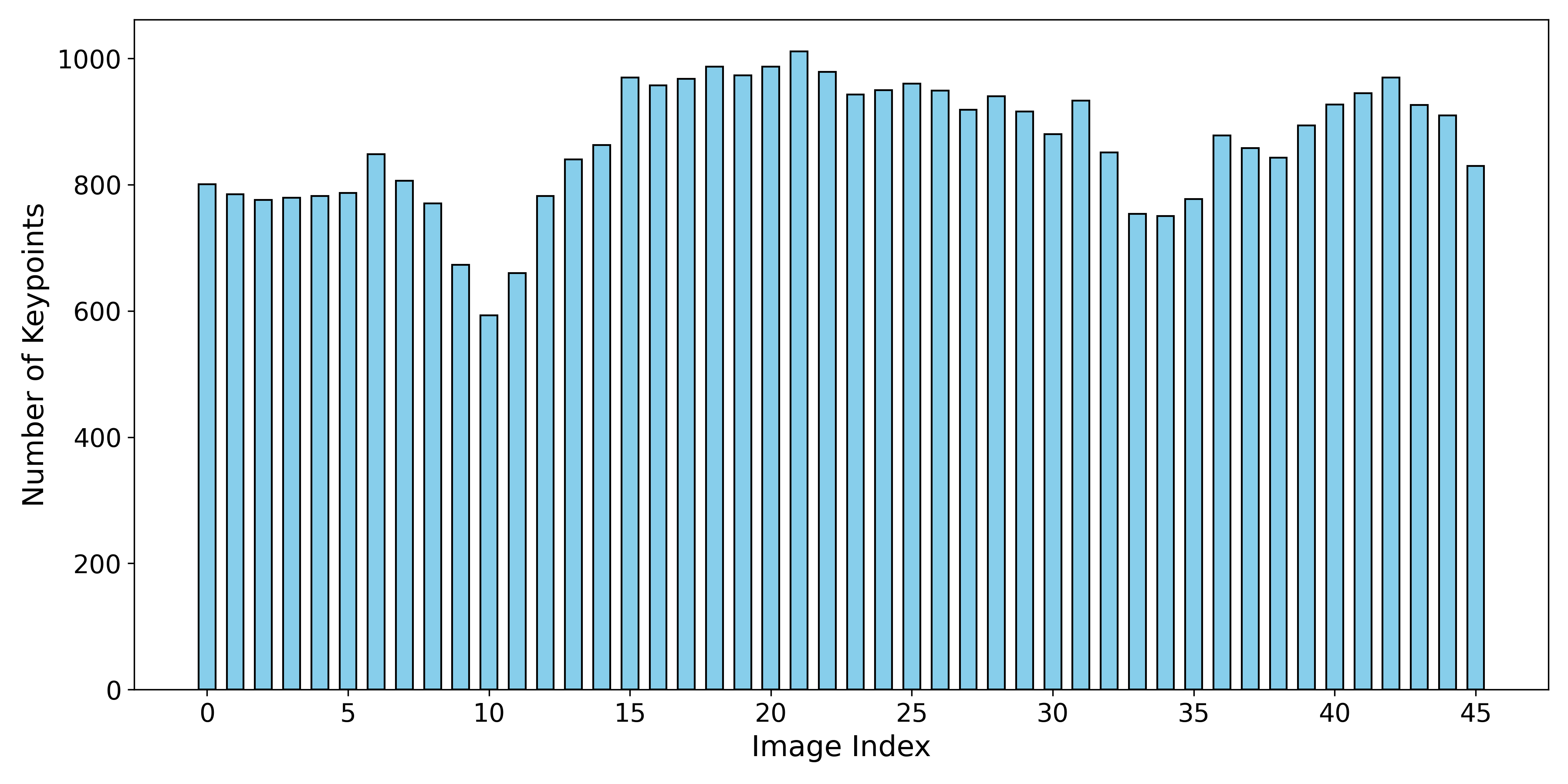}
        \caption{Number of Keypoints}
    \end{subfigure}

    \caption{Feature Detection: (a) Keypoints detected in Frame 0, (b) Keypoints detected in Frame 1, and (c) Histogram of \# features detected in dataset.}
    \label{fig:feature_detection}
\end{figure}

Keypoints were extracted using the SIFT algorithm with parameters selected to ensure a balance between detection robustness and computational efficiency. Three layers per octave were used in the scale space representation, which provided sufficient sampling across scales without introducing excessive computational overhead. The contrast threshold was set to 0.04 to discard low contrast keypoints that are more susceptible to noise and thus less stable across image variations. An edge response threshold $=10$ was used to filter out poorly localized features along edges, improving the distinctiveness of retained keypoints. The recommended Gaussian smoothing factor $=1.6$ was used for initial blurring \cite{Low04}.
These settings ensured robust detection of scale and rotation invariant features across the dataset. Fig.~\ref{fig:feature_detection}(a) and Fig.~\ref{fig:feature_detection}(b) show the keypoints detected in Frame~0 and Frame~1 respectively. Fig.~\ref{fig:feature_detection}(c) shows a histogram plot of the number of descriptors detected across the dataset. The average count was 867 per image.

\begin{figure}[t] 
    \centering
    \begin{subfigure}[b]{0.445\textwidth}
        \centering
        \includegraphics[width=\textwidth]{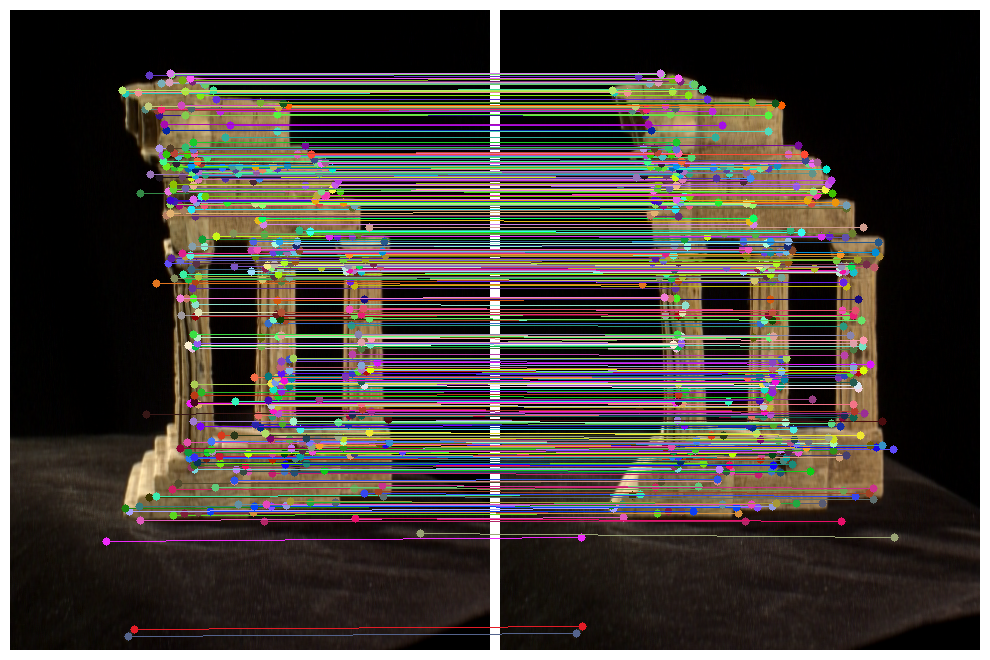}
        \caption{Feature Correspondences}
    \end{subfigure}

    \vspace{.5em}
    
    \begin{subfigure}[b]{0.48\textwidth}
        \centering
        \includegraphics[width=\textwidth, height=4.5cm]{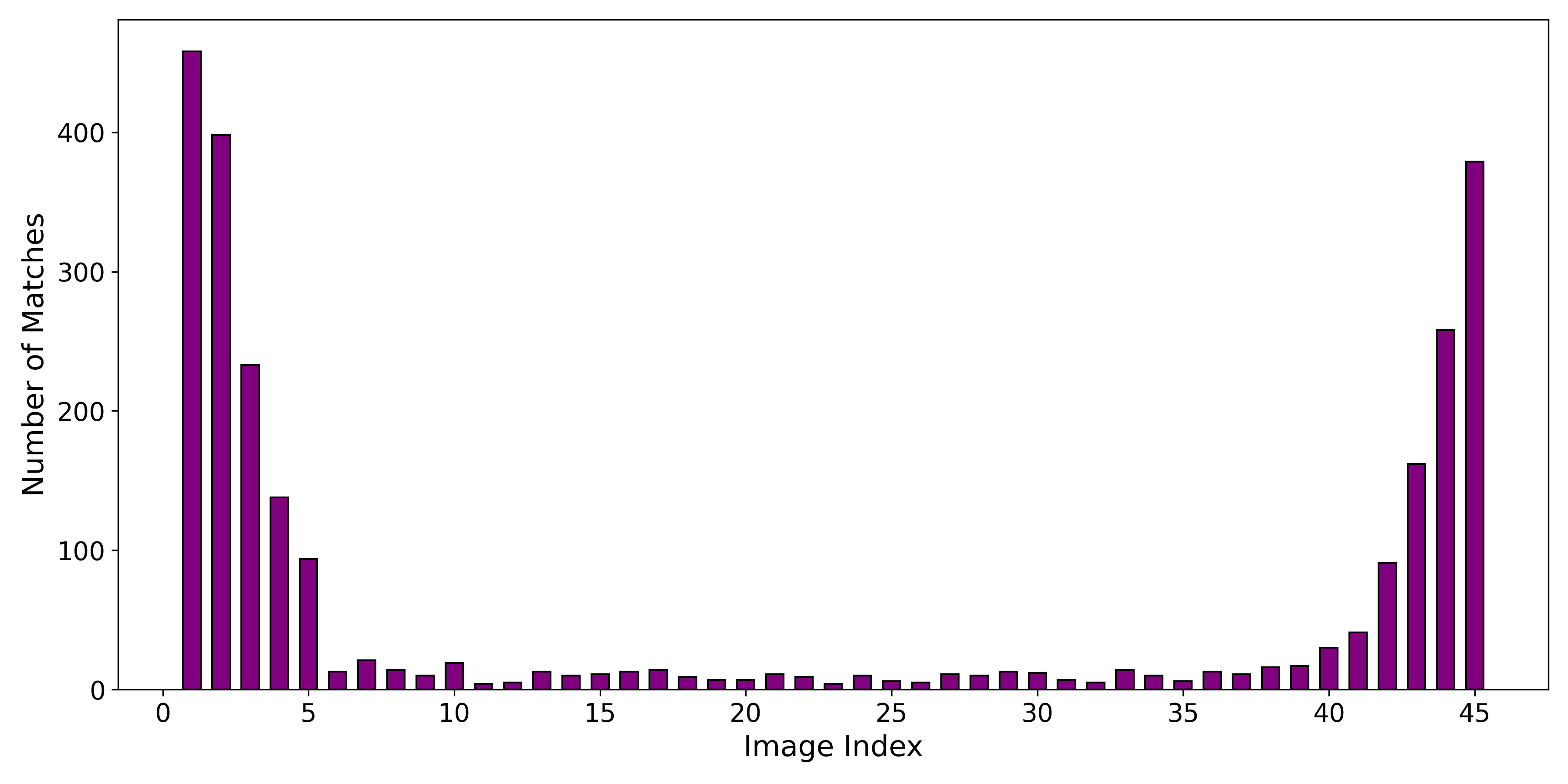}
        \caption{Number of Matched Keypoints}
    \end{subfigure}

    \vspace{.5em}
    
    \begin{subfigure}[b]{0.48\textwidth}
        \centering
        \includegraphics[width=\textwidth]{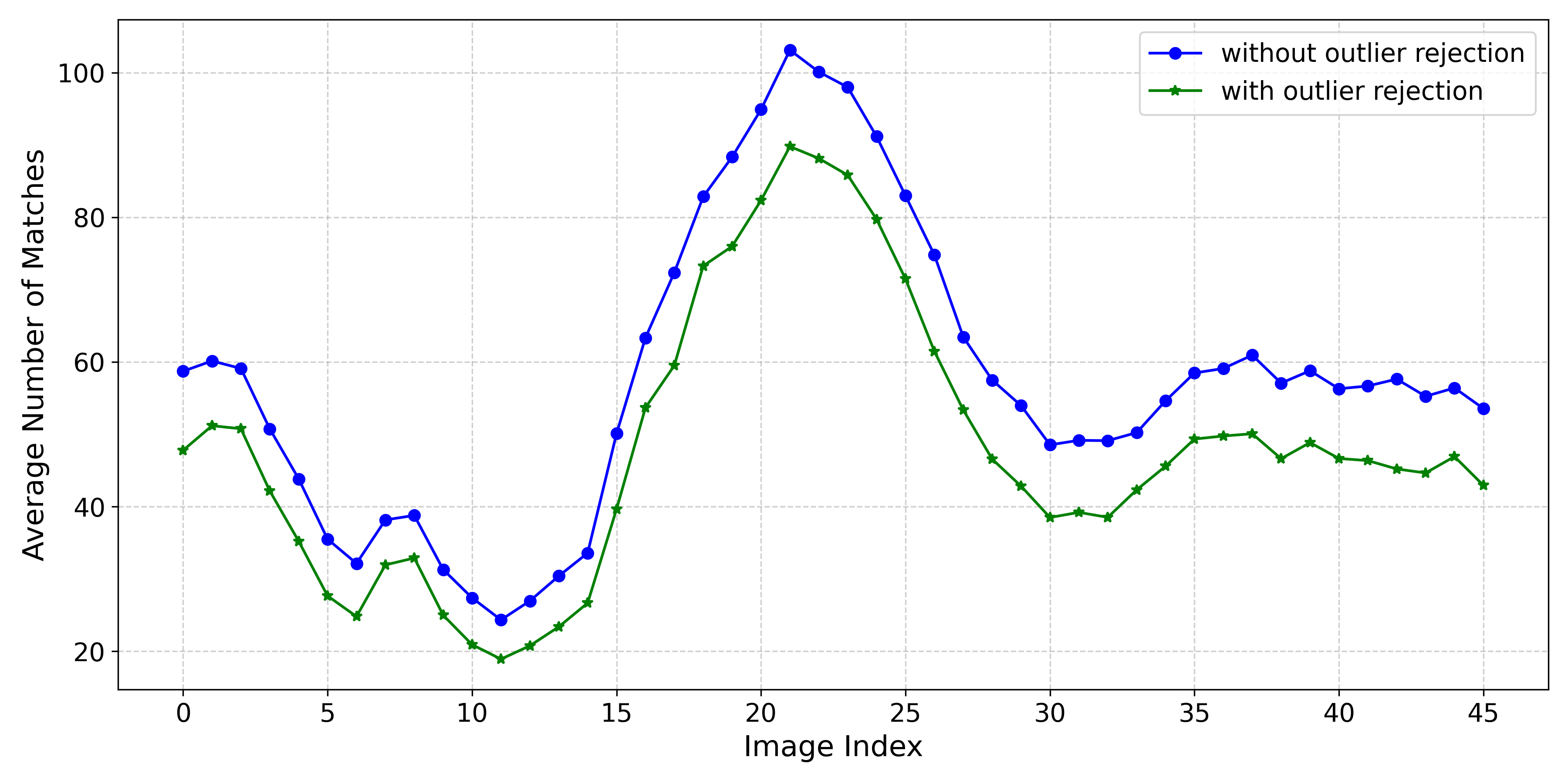}
        \caption{Average Matched Keypoints}
    \end{subfigure}

    \caption{Feature Matching: (a) Feature correspondences in Frame 0 and Frame 1, (b) \# keypoints in Frame 0 matched across dataset, and (c) Average \# matched keypoints across dataset before (blue) and after (green) outlier removal.}
    \label{fig:feature_matching}
\end{figure}

Features detected in image $i$ were matched across images $j>i$ for $i=1,2,...,N-1$. Initially, 458 putative matches were obtained based on descriptor similarity. After applying geometric verification using RANSAC to estimate the fundamental matrix, 439 matches were retained as inliers. Fig.~\ref{fig:feature_matching}(a) shows the matched features in Frame 0 and Frame 1. Fig.~\ref{fig:feature_matching}(b) shows the \# features of Frame 0 matched across the dataset. Higher match counts were observed for adjacent frames due to greater scene overlap. Fig.~\ref{fig:feature_matching}(c) shows the average \# features of an image matched across the dataset before (in blue) and after (in green) RANSAC-based outlier rejection. The average \# matches before outlier removal was \(\approx 58\) which was reduced to \(\approx 48\) after removing inconsistent correspondences.

Given that 801 and 785 features were originally detected in Frame 0 and Frame 1 respectively, the robustness of the features can be quantified using match ratios. Before outlier removal, we had \( \frac{458}{801} \approx 57.2\% \) matched features in Frame 0 and \( \frac{458}{785} \approx 58.3\% \) matched features in Frame 1. After outlier removal, the revised ratios were \( \frac{439}{801} \approx 54.8\% \) and \( \frac{439}{785} \approx 55.9\% \) respectively which indicate that over half of the initially detected features resulted in successful and geometrically consistent matches.

\begin{figure}
    \centering
    \includegraphics[width=0.85\linewidth]{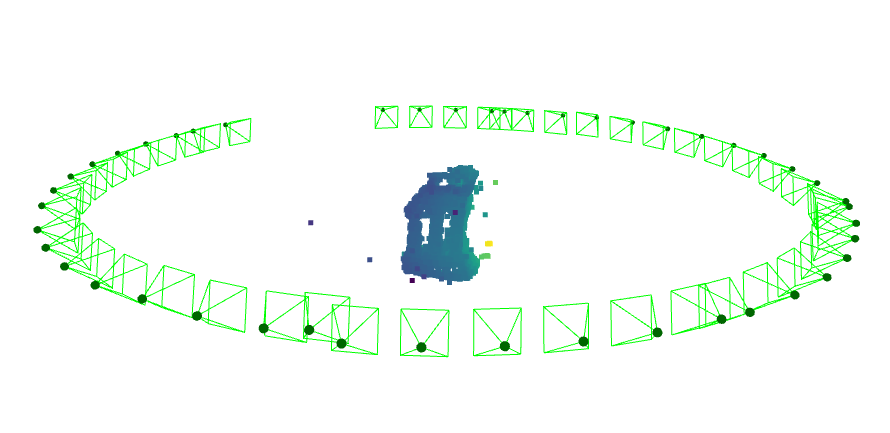}
    \caption{Estimated camera centers and sparse 3D point cloud.}
    \label{fig:camera_centers}
\end{figure}

Next, we performed camera pose estimation for each view. From the estimated projection matrix \eqref{e:M}, the camera center \( \mathbf{C}_i \) for image \( i \) was computed using \eqref{e:C2}. Fig.~\ref{fig:camera_centers} shows the estimated camera trajectory around the reconstructed scene. The green dots represent the estimated camera centers, while the green wireframe frustums represent the camera orientations and field of view which form a circular path around the structure, and is consistent with the acquisition setup. This spatial configuration validates the robustness of the motion estimation and provides a solid foundation for refining the 3D structure through bundle adjustment.

\begin{figure}
    \centering
    \includegraphics[width=0.48\textwidth]{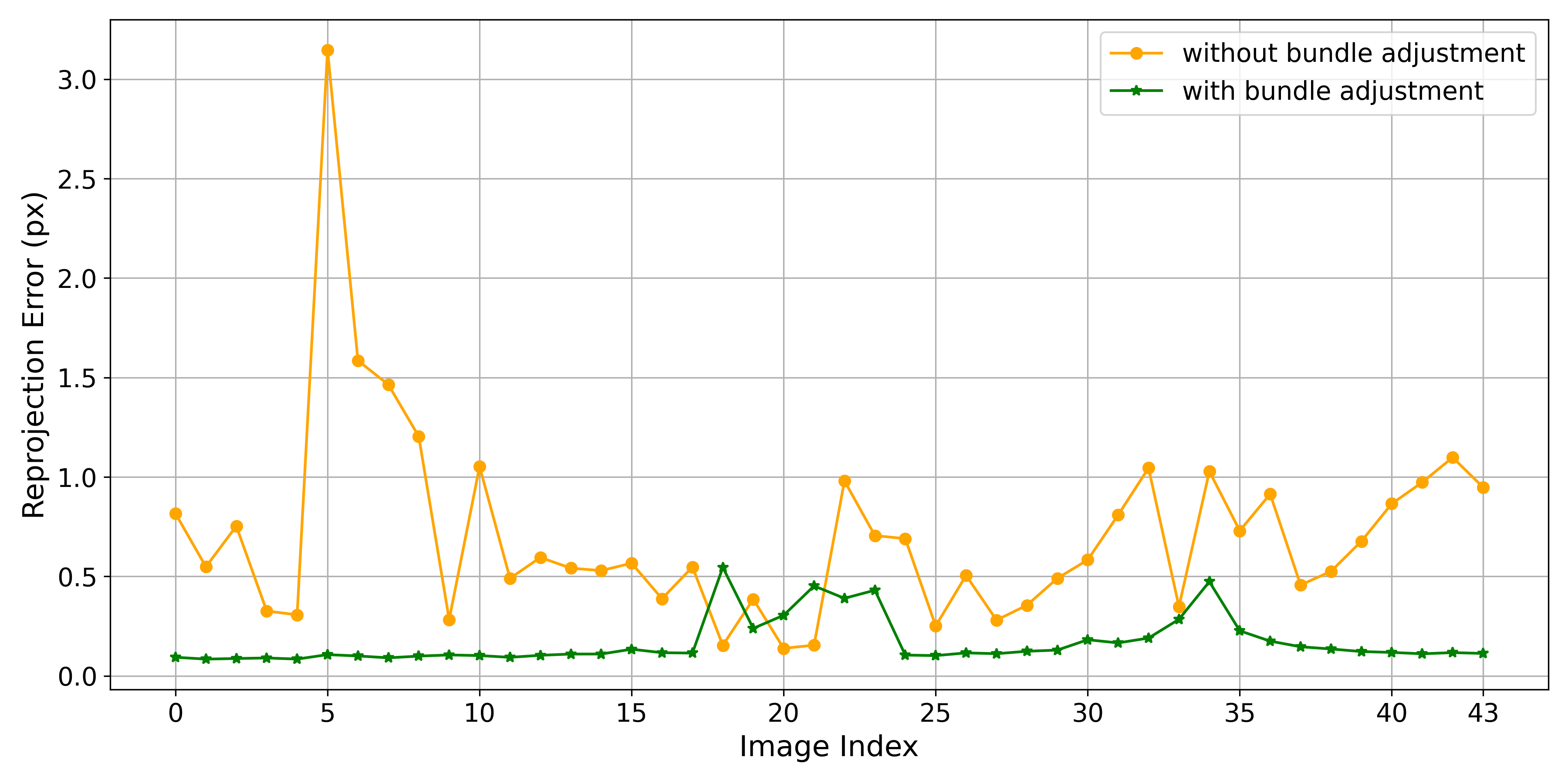}
    \caption{Reprojection Error: Before (orange) and after (green) bundle adjustment.}
    \label{fig:reproj_errors}
\end{figure}

To evaluate the accuracy of the estimated camera poses and 3D structure, we analyzed the reprojection error before and after bundle adjustment. Fig.~\ref{fig:reproj_errors} shows a comparison of the reprojection errors for each image in the sequence. The orange markers indicate the error before bundle adjustment, while the green markers indicate the errors after refinement. Clearly, bundle adjustment significantly reduced the reprojection error across most images, indicating improved alignment between the observed 2D feature locations and the reprojected 3D points. This confirms the effectiveness of the optimization in refining both camera parameters and 3D structure.

\begin{figure}[t]
    \centering
    \begin{subfigure}[b]{0.24\textwidth}
        \centering
        \includegraphics[width=\textwidth]{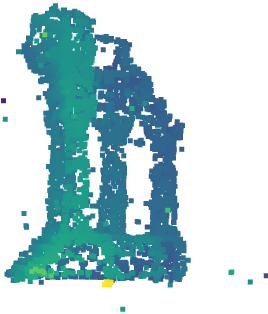}
        \caption{Front}
    \end{subfigure}
    \begin{subfigure}[b]{0.24\textwidth}
        \centering
        \includegraphics[width=\textwidth]{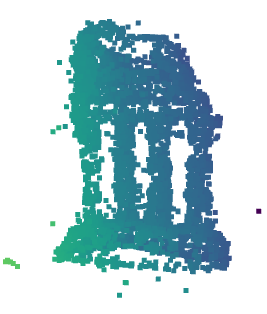}
        \caption{Right}
    \end{subfigure}
    
    \begin{subfigure}[b]{0.24\textwidth}
        \centering
        \includegraphics[width=\textwidth]{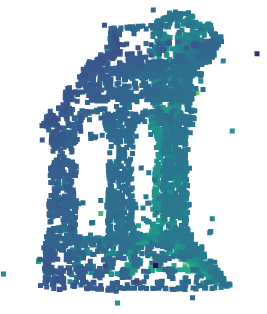}
        \caption{Back}
    \end{subfigure}
    \begin{subfigure}[b]{0.24\textwidth}
        \centering
        \includegraphics[width=\textwidth]{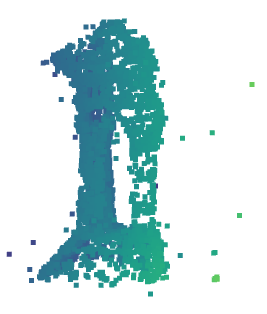}
        \caption{Left}
    \end{subfigure}
    
    \caption{Custom Incremental SfM: Four viewpoints.}
    \label{fig:3d_rec}
\end{figure}

The (sparse) 3D reconstruction generated by our custom incremental SfM pipeline was visualized from multiple viewpoints to assess the structural completeness and spatial consistency of the recovered scene geometry. Fig.~\ref{fig:3d_rec} shows the model from four perspectives: front, right, back, and left. The point cloud preserves key architectural features such as vertical columns and the stepped base, while maintaining overall consistency across views. Although the reconstruction is generally dense and coherent, some regions with poor texture or limited visibility exhibit missing patches, likely due to insufficient feature correspondences during matching.

\begin{figure}[t]
    \centering
    \begin{subfigure}[b]{0.24\textwidth}
        \centering
        \includegraphics[width=\textwidth]{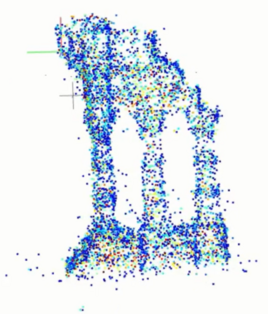}
        \caption{Front}
    \end{subfigure}
    \begin{subfigure}[b]{0.24\textwidth}
        \centering
        \includegraphics[width=\textwidth]{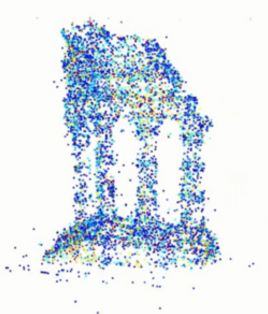}
        \caption{Right}
    \end{subfigure}
    
    \begin{subfigure}[b]{0.24\textwidth}
        \centering
        \includegraphics[width=\textwidth]{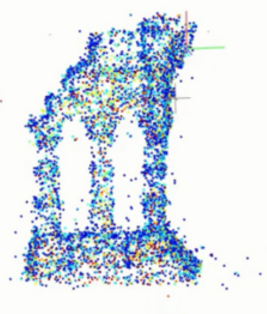}
        \caption{Back}
    \end{subfigure}
    \begin{subfigure}[b]{0.24\textwidth}
        \centering
        \includegraphics[width=\textwidth]{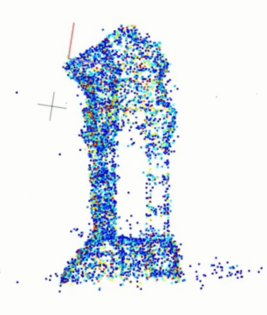}
        \caption{Left}
    \end{subfigure}
    
    \caption{COLMAP: Four viewpoints.}
    \label{fig:colmap}
\end{figure}

To benchmark our approach, 3D reconstruction was performed using COLMAP \cite{she2024refractive}. Fig.~\ref{fig:colmap} shows the four viewpoints. COLMAP successfully reconstructed the overall geometry of the scene without significant missing patches, particularly in regions where our pipeline shows gaps. However, the reconstruction exhibits slight noise near the top and along the outer edges, where dispersed points and fragmented structures are visible. These irregularities are likely due to minor feature mismatches or limited texture information at the periphery. Despite this, the core structure remains well-defined and coherent.

Both methods successfully reconstruct the global structure of the scene, but with notable trade-offs. The custom incremental SfM pipeline yields a denser reconstruction with finer architectural detail, while COLMAP ensures broader coverage and fewer missing areas. However, this comes at the cost of increased noise, especially near boundaries. These results demonstrate that our SfM pipeline delivers performance comparable to COLMAP, while offering enhanced modularity and flexibility for customized research and development.

\section{Conclusions}

In this paper, we presented a complete and robust pipeline for 3D reconstruction using incremental Structure from Motion (SfM). Keypoints were detected using the SIFT algorithm to ensure invariance to scale and rotation, and feature correspondences were established using  brute-force matching based on the Euclidean distance. Camera poses were incrementally estimated, and 3D points were reconstructed through triangulation of matched keypoints across multiple views. To evaluate the accuracy of the reconstruction, reprojection error was computed, confirming that the reconstructed points were geometrically consistent with the observed image features. Experimental results demonstrated that the proposed approach effectively recovered accurate camera trajectories and sparse 3D structure.
A comparison with COLMAP demonstrated that the proposed incremental SfM pipeline delivered comparable geometrically consistent reconstruction.

\bibliographystyle{IEEEtran}
\bibliography{IEEEabrv,refs}

\begin{thebibliography}{10}
\providecommand{\url}[1]{#1}
\csname url@samestyle\endcsname
\providecommand{\newblock}{\relax}
\providecommand{\bibinfo}[2]{#2}
\providecommand{\BIBentrySTDinterwordspacing}{\spaceskip=0pt\relax}
\providecommand{\BIBentryALTinterwordstretchfactor}{4}
\providecommand{\BIBentryALTinterwordspacing}{\spaceskip=\fontdimen2\font plus
\BIBentryALTinterwordstretchfactor\fontdimen3\font minus
  \fontdimen4\font\relax}
\providecommand{\BIBforeignlanguage}[2]{{%
\expandafter\ifx\csname l@#1\endcsname\relax
\typeout{** WARNING: IEEEtran.bst: No hyphenation pattern has been}%
\typeout{** loaded for the language `#1'. Using the pattern for}%
\typeout{** the default language instead.}%
\else
\language=\csname l@#1\endcsname
\fi
#2}}
\providecommand{\BIBdecl}{\relax}
\BIBdecl

\bibitem{HZi03}
R.~Hartley and A.~Zisserman, \emph{Multiple View Geometry in Computer
  Vision}.\hskip 1em plus 0.5em minus 0.4em\relax Cambridge University Press,
  2003.

\bibitem{szeliski2022computer}
R.~Szeliski, \emph{Computer Vision: Algorithms and Applications}.\hskip 1em
  plus 0.5em minus 0.4em\relax Springer Nature, 2022.

\bibitem{pierson2017deep}
H.~A. Pierson and M.~S. Gashler, ``Deep learning in robotics: A review of
  recent research,'' \emph{Advanced Robotics}, vol.~31, no.~16, pp. 821--835,
  2017.

\bibitem{lou2023slam}
L.~Lou, Y.~Li, Q.~Zhang, and H.~Wei, ``{SLAM} and 3{D} semantic reconstruction
  based on the fusion of lidar and monocular vision,'' \emph{Sensors}, vol.~23,
  no.~3, p. 1502, 2023.

\bibitem{xu2024real}
L.~Xu, Y.~Xu, Z.~Rao, and W.~Gao, ``Real-time 3d reconstruction for the
  conservation of the {G}reat {W}all’s cultural heritage using depth
  cameras,'' \emph{Sustainability}, vol.~16, no.~16, p. 7024, 2024.

\bibitem{wijayasinghe2024smart}
S.~Wijayasinghe and V.~Sachitra, ``Smart city development and improvement of
  quality of life in urban cities of {S}ri {L}anka: citizen-centric approach,''
  \emph{Journal of Global Responsibility}, 2024.

\bibitem{curless1996volumetric}
B.~Curless and M.~Levoy, ``A volumetric method for building complex models from
  range images,'' in \emph{Proc. 23rd Annual Conf. Computer Graphics and
  Interactive Techniques}, 1996, pp. 303--312.

\bibitem{SFr16}
J.~L. Schonberger and J.-M. Frahm, ``Structure-from-motion revisited,'' in
  \emph{Proc. IEEE Conf. Computer Vision and Pattern Recognition (CVPR)}, 2016,
  pp. 4104--4113.

\bibitem{FHe15}
Y.~Furukawa, C.~Hern{\'a}ndez \emph{et~al.}, ``Multi-view stereo: {A}
  tutorial,'' \emph{Foundations and trends{\textregistered} in Computer
  Graphics and Vision}, vol.~9, no. 1-2, pp. 1--148, 2015.

\bibitem{triggs1999bundle}
B.~Triggs, P.~F. McLauchlan, R.~I. Hartley, and A.~W. Fitzgibbon, ``Bundle
  adjustment--a modern synthesis,'' in \emph{Intl. Wrkshp. Vision
  Algorithms}.\hskip 1em plus 0.5em minus 0.4em\relax Springer, 1999, pp.
  298--372.

\bibitem{she2024refractive}
M.~She, F.~Seegr{\"a}ber, D.~Nakath, and K.~K{\"o}ser, ``Refractive {COLMAP}:
  refractive structure-from-motion revisited,'' in \emph{IEEE/RSJ Intl. Conf.
  Intelligent Robots and Systems (IROS)}, 2024, pp. 12\,816--12\,823.

\bibitem{chen2023adasfm}
Y.~Chen, Z.~Yu, S.~Song, T.~Yu, J.~Li, and G.~H. Lee, ``Ada{S}f{M}: From coarse
  global to fine incremental adaptive structure from motion,'' in \emph{IEEE
  Intl. Conf. Robotics and Automation (ICRA)}, 2023, pp. 2054--2061.

\bibitem{cui2023mcsfm}
H.~Cui, X.~Gao, and S.~Shen, ``{MCS}f{M}: multi-camera-based incremental
  structure-from-motion,'' \emph{IEEE Trans. Image Processing}, vol.~32, pp.
  6441--6456, 2023.

\bibitem{mateus2021incremental}
A.~Mateus, O.~Tahri, A.~P. Aguiar, P.~U. Lima, and P.~Miraldo, ``On incremental
  structure from motion using lines,'' \emph{IEEE Trans. Robotics}, vol.~38,
  no.~1, pp. 391--406, 2021.

\bibitem{ranftl2020towards}
R.~Ranftl, K.~Lasinger, D.~Hafner, K.~Schindler, and V.~Koltun, ``Towards
  robust monocular depth estimation: Mixing datasets for zero-shot
  cross-dataset transfer,'' \emph{IEEE Trans. Pattern Analysis and Machine
  Intelligence}, vol.~44, no.~3, pp. 1623--1637, 2020.

\bibitem{zhao2022monovit}
C.~Zhao, Y.~Zhang, M.~Poggi, F.~Tosi, X.~Guo, Z.~Zhu, G.~Huang, Y.~Tang, and
  S.~Mattoccia, ``Mono{V}i{T}: Self-supervised monocular depth estimation with
  a vision transformer,'' in \emph{Intl. Conf. 3D Vision (3DV)}, 2022, pp.
  668--678.

\bibitem{yao2018mvsnet}
Y.~Yao, Z.~Luo, S.~Li, T.~Fang, and L.~Quan, ``{MVSN}et: Depth inference for
  unstructured multi-view stereo,'' in \emph{Proc. European Conf. Computer
  Vision (ECCV)}, 2018, pp. 767--783.

\bibitem{zhang2020review}
H.-M. Zhang and B.~Dong, ``A review on deep learning in medical image
  reconstruction,'' \emph{Journal of the Operations Research Society of China},
  vol.~8, no.~2, pp. 311--340, 2020.

\bibitem{golub2013matrix}
G.~H. Golub and C.~F. Van~Loan, \emph{Matrix Computations}.\hskip 1em plus
  0.5em minus 0.4em\relax JHU Press, 2013.

\bibitem{Low04}
D.~G. Lowe, ``Distinctive image features from scale-invariant keypoints,''
  \emph{International Journal of Computer Vision}, vol.~60, pp. 91--110, 2004.

\bibitem{marr1980theory}
D.~Marr and E.~Hildreth, ``Theory of edge detection,'' \emph{Proc. R. Soc.
  Lond. Ser. B. Biol. Sci.}, vol. 207, no. 1167, pp. 187--217, 1980.

\bibitem{freeman1991design}
W.~T. Freeman and E.~H. Adelson, ``Design and use of steerable filters,''
  \emph{IEEE Trans. Pattern Analysis and Machine Intelligence}, vol.~13, no.~9,
  pp. 891--906, 1991.

\bibitem{MJR22}
J.~M. Mart{\'\i}nez-Otzeta, I.~Rodr{\'\i}guez-Moreno, I.~Mendialdua, and
  B.~Sierra, ``{RANSAC} for robotic applications: {A} survey,'' \emph{Sensors},
  vol.~23, no.~1, p. 327, 2022.

\bibitem{Har97}
R.~I. Hartley, ``In defense of the eight-point algorithm,'' \emph{IEEE Trans.
  Pattern Analysis and Machine Intelligence}, vol.~19, no.~6, pp. 580--593,
  1997.

\bibitem{nister2004efficient}
D.~Nist{\'e}r, ``An efficient solution to the five-point relative pose
  problem,'' \emph{IEEE Trans. Pattern Analysis and Machine Intelligence},
  vol.~26, no.~6, pp. 756--770, 2004.

\bibitem{NWr99}
J.~Nocedal and S.~J. Wright, \emph{Numerical Optimization}.\hskip 1em plus
  0.5em minus 0.4em\relax Springer, 1999.

\bibitem{lu2018review}
X.~X. Lu, ``A review of solutions for perspective-n-point problem in camera
  pose estimation,'' in \emph{Journal of Physics: Conference Series}, vol.
  1087, no.~5.\hskip 1em plus 0.5em minus 0.4em\relax IOP Publishing, 2018, p.
  052009.

\bibitem{lepetit2009ep}
V.~Lepetit, F.~Moreno-Noguer, and P.~Fua, ``E{P}n{P}: An accurate {O}(n)
  solution to the {P}n{P} problem,'' \emph{International Journal of Computer
  Vision}, vol.~81, pp. 155--166, 2009.

\end{thebibliography}

\end{document}